\documentclass{vgtc}                          

\ifpdf
  \pdfoutput=1\relax                   
  \usepackage{graphicx}                
  \DeclareGraphicsExtensions{.pdf,.png,.jpg,.jpeg} 
\else
  \usepackage{graphicx}                
  \DeclareGraphicsExtensions{.eps}     
\fi%

\graphicspath{{figures/}{pictures/}{images/}{./}} 

\usepackage{epstopdf}
\usepackage{microtype}                 
\PassOptionsToPackage{warn}{textcomp}  
\usepackage{textcomp}                  
\usepackage{mathptmx}                  
\usepackage{times}                     
\usepackage{cite}                      
\usepackage{tabu}                      
\usepackage{booktabs}
\usepackage{enumitem}
\usepackage{array}
\usepackage{multirow}
\usepackage{ragged2e}
\usepackage{fontawesome}

\usepackage{authblk} 

\usepackage{xcolor}

\onlineid{1448}

\vgtccategory{Research}

\vgtcinsertpkg

\begin{document}
\newcommand{\sysName}{NFTVis}
\title{{\sysName}: Visual Analysis of NFT Performance}




\author[1]{Fan Yan\thanks{e-mail: yanfan@zju.edu.cn}}
\affil[1]{\scriptsize State Key Lab of CAD\&CG, Zhejiang University}
\author[2]{Xumeng Wang\thanks{e-mail: wangxumeng@nankai.edu.cn}}
\affil[2]{\scriptsize College of Computer Science, Nankai University}
\author[1]{Ketian Mao\thanks{e-mail: 22151321@zju.edu.cn}}
\author[1]{Wei Zhang\thanks{e-mail: zw\_yixian@zju.edu.cn}}
\author[1, 3]{Wei Chen\thanks{e-mail: chenvis@zju.edu.cn (corresponding author)}}
\affil[3]{Laboratory of Art and Archaeology Image (Zhejiang University),
Ministry of Education, China}

\teaser{
  \centering
  \includegraphics[width=\linewidth]{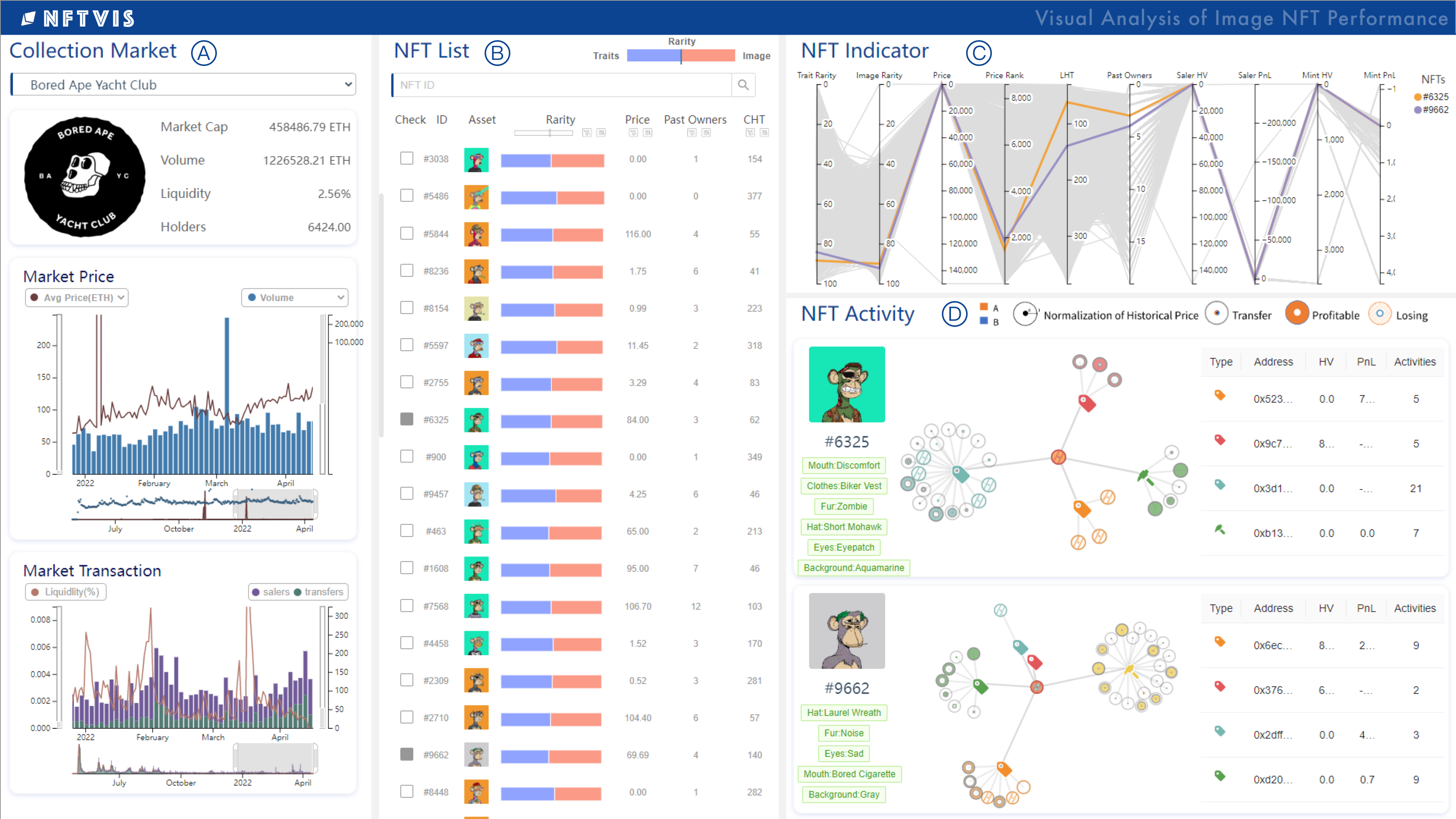}
  \caption{\sysName{} contains four views. A) The collection market view contains the market summary, market price chart, and market transaction chart. B) The NFT List view presents the basic information of NFTs and displays their traits rarity and image rarity. C) The NFT indicator view displays the NFT indicators of collection using parallel coordinates. D) The NFT activity view displays the transactions of NFTs.}
  \label{fig:system_design}
}

\abstract{A non-fungible token (NFT) is a data unit stored on the blockchain. Nowadays, more and more investors and collectors (NFT traders), who participate in transactions of NFTs, have an urgent need to assess the performance of NFTs.
However, there are two challenges for NFT traders when analyzing the performance of NFT.
First, the current rarity models have flaws and are sometimes not convincing.
In addition, NFT performance is dependent on multiple factors, such as images (high-dimensional data), history transactions (network), and market evolution (time series). It is difficult to take comprehensive consideration and analyze NFT performance efficiently.
To address these challenges, we propose \sysName, a visual analysis system that facilitates assessing individual NFT performance. 
A new NFT rarity model is proposed to quantify NFTs with images. Four well-coordinated views are designed to represent the various factors affecting the performance of the NFT.
Finally, we evaluate the usefulness and effectiveness of our system using two case studies and user studies.
} 


\CCScatlist{
  \CCScatTwelve{Non-fungible token}{Finance visualization}{Rarity model}{}
}





\firstsection{Introduction}
\maketitle
A non-fungible token (NFT) is a data unit that is stored on the blockchain.
In contrast to other cryptocurrencies, such as Bitcoin and Dogecoin, the token in an NFT is uniquely identifiable~\cite{wang2021non}.
There are many types of data assets stored in NFTs, including images, audio, videos, text, etc. The most popular NFTs are the NFTs with images~\cite{phillips202110}, so the NFT mentioned by most NFT traders usually is the NFT with images ~\cite{nadini2021mapping}.

A growing number of artists are storing their artworks in NFTs and selling them in the marketplace, providing NFTs with collectible value~\cite{singh2021distributed, whitaker2019art,franceschet2021sentiment}.
Nowadays, the NFT market has become a major marketplace for artwork exchanges~\cite{trautman2021virtual}.
By May 1, 2022, the NFT market for 2022 had received 37 billion USD, nearly catching up to the NFT market for 2021~\cite{fortune20220505}.
The NFT market is growing fast, and experts predict that NFT sales will soar in the next decade~\cite{jdsupra20220428}.
The booming market has attracted a large number of NFTs.

NFT traders often transact NFTs through NFT marketplaces.
However, there are two main challenges for NFT traders in analyzing the performance of interested NFTs.
First, the current rarity models~\cite{nftgo_rarity, rarity_tools_model}\footnote{The rarity of an NFT is defined by its differences from other NFTs.} do not convince NFT traders in some cases.
For example, a collection\footnote{The collection is a category that is set up based on the artists' creative ideas.} contains an equal number of brown-eyed, black-eyed, and white-eyed monkeys. The current rarity models based on traits\footnote{Traits are text labels that describe the image of NFTs.} would assign the same rarity to these three types of monkeys.
However, white is more visually distinguishable than black and brown for NFT traders.
NFT traders may consider white-eyed monkeys as rarer than others.
Therefore, a new rarity model that can refer to the image is required.
In addition, the performance of NFT data is multimodal, consisting of time-series data (market evolution), network data (transactions), and high-dimensional data (images of NFTs).
Current NFT marketplaces~\cite{opensea, cryptoart, azuki, bayc} display the performance of NFTs using basic charts and tables, which are not enough for NFT traders to efficiently analyze the performance of NFTs due to the complexity of the data.

To address these challenges, 
We designed a rarity model that quantifies the rarity of an NFT based on the difference between its image and the image of each other NFT in the same collection.
Furthermore, we proposed \sysName, a visual analysis system, to assist NFT traders in analyzing the performance of interested NFTs.
The system provides comprehensive views which facilitate the analysis of NFT performance from multiple perspectives.
The collection market view (\autoref{fig:system_design}A) helps NFT traders understand the temporal evolution of the collection's market. 
The NFT list view (\autoref{fig:system_design}B) allows NFT traders to select interested NFTs based on our rarity model. 
The NFT indicator view (\autoref{fig:system_design}C) enables NFT traders to assess the performance of NFT indicators on the collection market. 
The NFT activity view (\autoref{fig:system_design}D) uses a new visual design of the trading graph network to depict the trading patterns of each NFT trader in the target NFT.
To evaluate the usefulness and effectiveness of \sysName, we present two cases and collect feedback from user studies.

The main contributions of our work are as follows:
\begin{itemize}[nosep]
\item To the best of our knowledge, our work is the first visual analysis system that allows NFT traders to explore and analyze individual NFT.

\item We propose an image rarity model that quantifies the rarity of an NFT based on the difference between its image and the images of other NFTs in the same collection.

\item We conduct two cases to demonstrate our system.
The evaluation results of user studies prove the usefulness and effectiveness of our system.
\end{itemize}

\section{Related Work}
In this section, we introduce previous research on NFTs and visualization of financial investment.
\begin{figure*}[htb]
    \centering 
    \includegraphics[width=\linewidth]{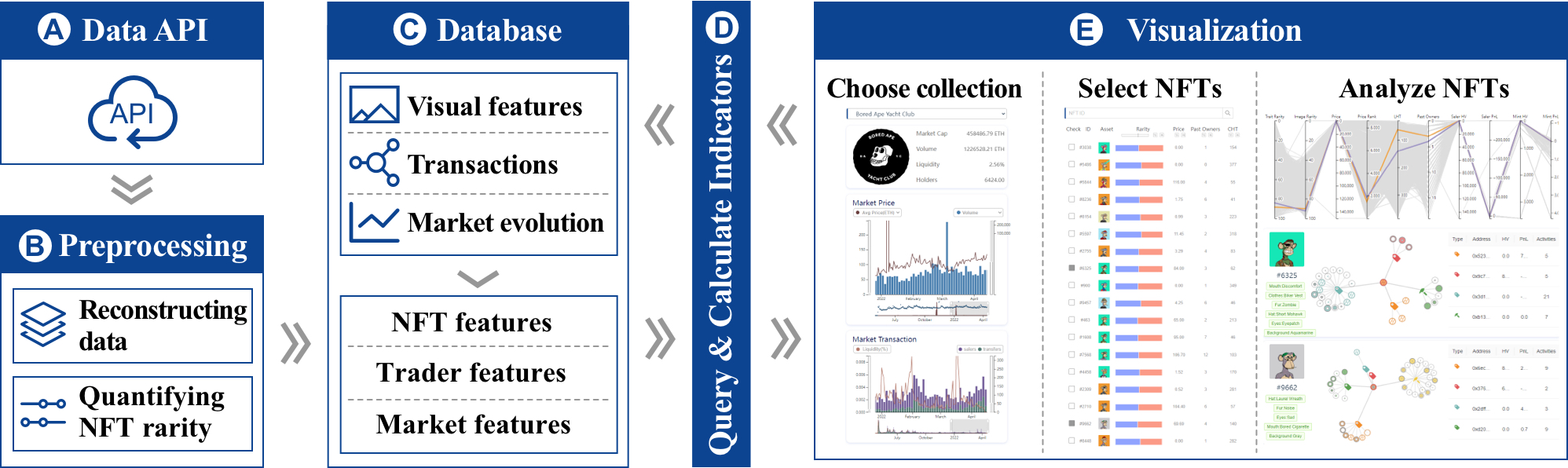}
    \caption{The system overview of \sysName, which consists of five modules. A) The data API module. B) The preprocessing module. C) The database module. D) The data service module. E) The visualization module.}
    \label{fig:system_overview}
\end{figure*}
\subsection{Finance Research of NFTs}
As the NFT market grows, more and more research is being done on NFTs. Researchers analyzed the NFT market mainly through three approaches: traditional financial models, machine learning, and visualizations.

Traditional financial models are leveraged to mine patterns in NFT data. Dowling~\cite{dowling2022non} and Ante~\cite{ante2021non2} used a spillover index, wavelet coherence, and co-integrated vector auto regressive (VAR) models to reveal a weak correlation between NFT and other cryptocurrency markets. Umar et al.~\cite{umar2022covid} 
used the wavelet approach to study coherence between return of NFTs and major assets.
Maouchi et al.~\cite{maouchi2021understanding} used statistical models to find an increase in NFT bubbles during COVID-19. Dowling et al.~\cite{dowling2022fertile} first priced virtual real estate of NFTs by statistical methods.
In addition to financial models, machine learning was also used to study NFTs. Franceschet et al.~\cite{franceschet2021hits} used Kleinberg's authority/hub HITS (Hyperlink-Induced Topic Search) method~\cite{kleinberg1999authoritative} to rate NFTs' artists and collectors. 
Kapoor et al.~\cite{kapoor2022tweetboost} fed social media data to machine learning models that could predict NFTs' asset value.
Other research used visualizations to understand NFT market data. Vasan et al.~\cite{vasan2022quantifying} built the NFT visual network to characterize artist-collector relationships. Nadini et al.~\cite{nadini2021mapping} visualized the overview of the NFT market to identify specific phenomena.

However, all previous research focuses on exploring the NFT market's insights.
There is no research available for NFT traders to assess the performance of individual NFTs in the collection.

\subsection{Visualization of Financial Investment}
There has been numerous research on financial investment visualization. We categorize these research according to their data into three groups: portfolio data, trade data, and other cryptocurrency data. In addition, we will discuss the differences between these data and the NFT data separately.

Portfolio data consists of equities, mutual funds, bonds, etc. The research of the visualization of portfolio data has a lengthy history. Wattenberg~\cite{wattenberg1999visualizing} improved Shneiderman's treemap~\cite{shneiderman1992tree} to show stock data in the market in 1999 by using both hierarchy and similarity information. Dwyer et al.~\cite{dwyer2002visualising, dwyer2004visualising} proposed a 3D design to visualize asset changes in portfolios. Keim et al.~\cite{keim2006spectral} designed a novel Growth Matrix visualization to analyze assets. Yue et al.~\cite{yue2019sportfolio, yue2021iquant} presented a visual analysis system for visualizing the performance of portfolio investment factors.
Portfolio data focuses on the market performance of products.
In contrast, NFT data focuses more on the patterns of transactions and traders.

Trade data consists of transaction information and numerous types of trade objects, such as companies and countries. Semulders and Heijs~\cite{smeulders2005interactive} used a 3D visualization to depict high-dimensional marketing data.
To imitate the real supply chain, Siddiqui et al.~\cite{siddiqui2008supply} developed an educational tool based on a scenario. Goh et al.~\cite{goh2013riskvis} created a visualization for an end-to-end supply chain to inspect potential risks. Basole et al.~\cite{basole2017visualization} used supply chain network visualization to uncover clusters, patterns, trends, and outliers.
Trade data visualizations focus on large-scale data, from which users may ignore features of individuals (e.g., traders, trade objects). Individual NFT visualizations have a small scale of trade objects. Each trade object has complex but essential features. It is necessary for NFT traders to observe such features and analyze the patterns of transactions and traders.

In terms of other cryptocurrencies' data visualization, Fleder et al.~\cite{fleder2015bitcoin} presented a transaction-graph-annotation system to explore the level of anonymity. McGinn et al.~\cite{mcginn2016visualizing} developed a large graph visualization of Bitcoin transactions to provide domain experts with useful insight. Yue et al.~\cite{yue2018bitextract} proposed a visual analytics approach for investigating and comparing the overview of several Bitcoin exchanges. Xia et al.~\cite{xia2020supoolvisor} developed a new visualization of cryptocurrency to supervise mining pools and de-anonymization by visual reasoning.
In the scenarios of the above studies, cryptocurrencies have the same monetary value. However, each NFT has a unique token and a distinct monetary value. It is necessary to analyze each NFT individually.

In addition, traders and transactions of portfolio data and trade data are often recorded in large corporations. For business information security, few corporations are willing to make their data available for open-source research.
And there is limited visualization research on traders and transactions~\cite{ko2016survey}. 
In contrast, NFT data is traceable on the blockchain network, which is accessible to the public. As a result, NFT data support exploring the patterns of transactions and traders.
In comparison to other cryptocurrencies, the performance and prices of each NFT vary.
Individual analysis of the NFT's performance is required.

\section{Background}
In this section, we introduce the data abstract about NFT and the tasks that need to be considered when analyzing NFT performance.

\subsection{Technical Indicators}
NFT traders assess the performance of NFTs based on NFT features, NFT transactions, and NFT collection.

The NFT features consist of price, images, and traits. Some research institutes have proposed an indicator called rarity, which is correlated with their price~\cite{chohan2021non}. Usually, the rarer the NFT, the higher its price~\cite{nadini2021mapping}.

In addition to the NFT features, NFT traders also analyze the performance of NFT transactions~\cite{franceschet2021hits,vasan2022quantifying,nadini2021mapping}.
The number of past owners, the current hold time (CHT), and the longest hold time (LHT) are used to assess individual NFTs.
The holding value (HV)\footnote{The holding value is the sum of the trader's NFT price.}, Profit and Loss (PnL)\footnote{The PnL is the sum of a trader's realized profit and unrealized profit.}~\cite{pantz2013pnl,lipton2021automated}, and the number of activities\footnote{The history transactions of NFTs are saved in an attribute called activity.} are used to assess NFT traders.

The NFTs collections are the categories that are set up based on the artists' creative ideas. The number of NFTs in the collection is not fixed. Large collections, such as Meebits and Bored Ape Yacht Club, have tens of thousands of NFTs. Existing research indicates that a good collection is continuously appealing to NFT traders~\cite{nadini2021mapping, ante2021non}. NFT traders typically assess the performance of a collection using the market cap, average price, floor price, volume, liquidity, and whale\&normal\footnote{
Some NFT traders who hold significant amounts of NFTs are referred to as whale traders. Other traders are referred to as normal traders.
} sale.
The market cap is the sum of the NFT prices in the collection, and the volume is the total volume of NFT sales during the statistical period.
The liquidity of the collection is calculated by $\frac{Sales}{\# NFTs} \times 100\%$.

\subsection{Requirements Analysis}

Over the past six months, we have collaborated closely with two NFT researchers ($E_1$ and $E_2$) to assist NFT traders in analyzing the performance of individual NFTs.
We conducted weekly online meetings with our collaborators. Each meeting lasted around one hour. They shared their NFT investment experience and the information service platforms (such as OpenSea~\cite{opensea}, Element~\cite{element}, OKLink~\cite{OKLink}, and NFTGo~\cite{nftgo}) that they use to select NFTs. 
In addition, we also gathered the NFT investment experience shared on the Internet. 
Based on the descriptions of these investment experiences, we have come up with four main requirements that NFT traders use to judge the performance of NFT.

\begin{enumerate}[label=\textbf{R{\arabic*}}, nolistsep]
    \setcounter{enumi}{0}
    \item \textit{Visualize the evolution of collection markets.}
    According to our researchers, the more active the collection market is, the more possible it is to attract additional NFT traders. And the majority of NFT traders frequently trade NFTs in the same collection market~\cite{nadini2021mapping}. Consequently, it is essential for NFT traders to have a comprehensive understanding of the collection market.

    \item \textit{Quantify NFT rarity with image and traits.}
    The current rarity models~\cite{nftgo_rarity, rarity_tools_model} calculate NFT rarity based on traits. They do not take the performance of the NFT images into account.
    So the current rarity models do not persuade NFT traders when they encounter similar traits with clearly distinguishable images.
    Therefore, there is a need for an NFT rarity model that considers both image and traits.
    
    \item \textit{Analyze the indicators of individual NFT with the collection.}
    The researchers note that evaluating each NFT's indicators should not be focused on the high or low value of the indicators but should also take into account other NFT indications from the same collection.
    However, many NFT information service platforms~\cite{opensea,nftgo,rarity.tools} only present the values of selected NFT indicators, making it difficult for NFT traders to assess the performance of individual NFTs.

    \item \textit{Reveal the patterns of NFT transaction.}
    Existing research~\cite{vidal2022new,parham2022non,pinto2022nft,pelechrinis2022spotting,sharma2022s} indicates that the NFT market contains anomalous transactions. These anomalous transactions may be speculation,  money laundering, etc. They artificially increase the cost of NFTs and disrupt the normal market for NFTs. Our experts have thought that if we were to disclose these patterns to NFT traders, then those traders would be in a better position to assess the performance of NFT.
\end{enumerate}
\section{SYSTEM OVERVIEW}

Our system is a single-page web application that assists NFT traders in analyzing NFT performance. The system consists of the five modules illustrated in \autoref{fig:system_overview}: (a) data API module, (b) preprocessing module, (c) database module, (d) data service module, and (e) visualization module. The data API module gathers the NFTs of collections from NFT information service platforms. Then the preprocessing module will reconstruct the data into NFT features, trader features, and market features. On the basis of NFT features, the rarity model (Section 5) will calculate the rarity of images and traits. After preprocessing, the database module will store all of the features. When NFT traders use the system for analysis, the service module will query the target NFTs' features and calculate real-time NFT indicators in the data service module.

In the visualization module, traders can assess the performance of NFTs. First, they can select a collection from the collection market view. Then, the collection market view will show an overview of the market using the market price chart and market transaction chart (\textbf{R1}). Simultaneously, the system will query the NFTs of the selected collection in the database module and calculate their indicators (Section 3.2) to the NFT list view. 
NFT traders can explore interested NFTs by ranking and filtering rarity and price in the NFT list view (\textbf{R2}). When NFT traders select an interested NFT, the NFT indicator view allows them to analyze the performance of the NFT's indicators with the collection (\textbf{R3}). In addition, NFT traders can analyze the transactional pattern of the selected NFT in the NFT activity view (\textbf{R4}).

Our system contains three well-known and sizeable Collections: CyptoPunks~\cite{cryptopunks}, Bored Ape Yacht Club~\cite{bayc}, and Azuki~\cite{azuki}. Each of them has tens of thousands of NFTs.
The CyptoPunks includes transactions from 23 July 2017 to 1 May 2022. The Bored Ape Yacht Club includes transactions from 1 May 2021 to 1 May 2022. And the Azuki contains transactions from 12 January 2022 to 1 May 2022.
The data API module, preprocessing module, and data service module are implemented with the Python and Flask backend framework. The database module is implemented with MongoDB. The visualization module is implemented with Vue.js and D3.js~\cite{bostock2011d3}.
Our system is hosted on a computer with an Intel i7-7700 CPU and 64GB of RAM.

\section{Date Model}
In this section, we introduce the rarity models.
The rarity model can tell NFT traders how rare their NFTs are. 
The rarity of an NFT is primarily reflected by two factors: 
1. the similarity of the NFT to other NFTs in the collection. The lower the similarity, the higher the rarity; 
2. The number of other NFTs with low similarity in the same collection. the larger the number, the higher the rarity of the NFT;
We will calculate the rarity of NFT in terms of traits and images separately.

\subsection{Trait rarity model}
We utilize the NFTGO rarity model to determine the rarity of traits, which is regarded as the best model for calculating trait rarity~\cite{explaining_nft_rarity}. It employs the Jaccard distance to assess the similarity between each trait.
The trait rarity function is
\autoref{eq:trait_rarity}. 
\begin{equation}
    T_z=\frac{1}{m}\sum_{i=1}^{m}(1-JD(z,m_i)),
    \label{eq:trait_rarity}
\end{equation}
where $T_z$ is the trait rarity of NFT $z$ and $m$ is the number of NFTs in a collection.
Jaccard distance (JD) is an algorithm to calculate traits in \autoref{eq:jd}.  
\begin{equation}
    JD(Tx,Ty) = \frac{|Tx \cap Ty|}{|Tx|+|Ty|-|Tx \cap Ty|}
    \label{eq:jd}
\end{equation}
$Tx$ and $Ty$ are the traits of NFT $x$ and NFT $y$.

\subsection{Image rarity model}
In addition, we designed an image rarity model.
The model first quantifies the difference between the NFT image and the image of each other NFT in the same collection. Then, the image rarity of the NFT is calculated by the average value of the differences.
Our image rarity can be obtained by the following three steps:

To quantify differences, we calculate the NFT visual features for the RGB color channels using scale-invariant feature transform (SIFT)~\cite{lowe1999object}. 
First, we search across all possible scales to detect potential feature points in the one channel of each NFT image. An image's scale space is defined as the function $L(x,y,\sigma)$ obtained by convolving the available-scale Gaussian $G(x,y,\sigma)$ with the input image $I(x,y)$ in \autoref{eq:spatial_sequence}.
\begin{equation}
    L(x,y,\sigma) = G(x,y,\sigma) \bigotimes I(x,y) \label{eq:spatial_sequence}
\end{equation} 
\begin{equation}
    G(x,y,\sigma) = \frac{1}{2\pi\sigma^2}e^{-\frac{(x^2+y^2)}{2\sigma^2}}
\label{eq:gaussian}
\end{equation}
Where $\sigma$ is the parameter of scale-space. $\bigotimes$ is the convolution operator.
Next, we use the difference-of-Gaussian (DoG) function to detect stable feature point locations in scale space. The DoG function $D(x,y,\sigma)$ is computed using the difference between two adjacent Gaussian images in the scale space in \autoref{eq:dog}.
\begin{equation}
    D(x,y,\sigma) = L(x,y,k\sigma) - L(x,y,\sigma) \label{eq:dog}
\end{equation}
Where $k$ is a constant multiplicative factor.
After that, we calculate the local extrema. We compare each feature point with $8$ neighboring pixels in the same scale and $2 \times 9$ neighboring pixels in the adjacent scales. If the central pixel is the maximum or minimum value of other neighboring pixels, then the central pixel is the local extrema.
Then, we define the dominant orientation of each feature point. Each feature point's dominant orientation is the highest peak of the image region's gradient orientation histogram. We calculate the image gradient magnitude $F(x,y)$ and orientation $H(x,y)$ based on the pixel difference through \autoref{eq:magnitude} and \autoref{eq:orientation}.
\begin{equation}
    F(x,y) = \sqrt{(L(x+1,y)-L(x-1,y))^2+(L(x,y+1)-L(x,y-1))^2} \label{eq:magnitude}
\end{equation}
\begin{equation}
    H(x,y) = tan^{-1}{\left(\frac{L(x,y+1)-L(x,y-1)}{L(x+1,y)-L(x-1,y)}\right)} \label{eq:orientation}
\end{equation}
Finally, we obtained a 128-dimensional NFT visual feature in one channel of an NFT image by calculating the gradient magnitudes and orientations of the local image in the 16x16 region around the feature points. Taking into account the effects of illumination changes, the NFT visual feature was normalized.

In the second step, we calculated the difference between NFT images in the same collections for the RGB color channels based on approximate algorithm~\cite{lowe2004distinctive}. We first paired NFT visual features using the Best-Bin-First(BBF)~\cite{beis1997shape} and calculated the Euclidean distance between each pair. Then, we compare the ratio between the minimum distance and the subsequent minimum distance to the threshold. If the ratio is less than the threshold, it is considered a match; otherwise, it is considered a no-match. The default threshold is 0.8. Finally, we calculated the difference between NFT images by the ratio of no matches.

In the last step, we calculated the image rarity of each NFT by averaging the difference for the RGB color channels between each NFT image and the other NFT images in the same collection, as shown in \autoref{eq:image_rarity}.
\begin{equation}
    IR_Z=\frac{1}{3M}\sum_{i=1}^{M}{\left[Dif(Z,M_i)^R+Dif(Z,M_i)^G+Dif(Z,M_i)^B\right]} \label{eq:image_rarity}
\end{equation}
Where $IR_Z$ is the image rarity of NFT $Z$. $M$ is the number of NFTs in a collection. $Dif$ is the difference between NFT images in the same collections. $R$, $G$, and $B$ are the color channels of NFT images.


\section{VISUALIZATION}
In this section, we will detail the purpose of each view and the properties that are displayed.
\subsection{Collection Market View}
The collection market view (\autoref{fig:system_design}A) assists NFT traders in selecting the collection. The view illustrates the evolution of the collection market (\textbf{R1}) by displaying the trend of market indicators. The drop-down menu allows NFT traders to choose the interested collection.
Below the drop-down menu is a summary of the collection's performance, including the image, market cap, volume, liquidity, and holders. When NFT traders hover over the collection's image, a tooltip will display the collection's description and official URL.

We provide the market price chart and market transaction chart to reveal the collection market's performance in detail. The market price chart displays indicators that are correlated with price. We use a dual-axis chart to analyze the relationship between the prices of NFTs and their sales. The left y-axis indicators include market cap, average price, and floor price.
Since these technical indicators are used to observe trends, we plot them using a line chart.
The right y-axis indicators include volume and whale \& normal sales. We use bar charts to depict volume, which is a statistical indicator. And we use scatterplots, which are discrete transaction records, to depict whale's normal sales.
We use the logarithmic scale on the right y-axis because there can be a million-fold difference between the minimum and maximum values.

The market transaction chart displays all of the transactions in the collection. The view also includes dual-axis. The left y-axis indicator is liquidity. We use a linear chart to observe the trend of market liquidity. The right y-axis indicators are the number of sales and transfers. We plot both indicators simultaneously using stacked bar charts. The length of the bar represents the number of daily transactions in the collection, and traders can investigate individual indicators by clicking on their corresponding legends.

To facilitate better analysis, we use a custom zoom layout for both the x- and y-axes. The market price and market transaction charts will show the same interaction period because their x-axes are the same.

\subsection{NFT List View}
The NFT list view (\autoref{fig:system_design}B) focuses on orderly recommending NFTs of interest to NFT traders. 
The view contains basic information about the NFT, such as id, image, rarity, price, and so on.
NFT traders can select NFTs from the list view to analyze and assess their performance in the NFT Indicator view and the NFT Activity view.

The  NFT list view is composed of three primary components: the search bar, the rarity weighting slider, and the list. The search bar allows NFT traders to look up specific NFTs by their ids. The rarity weighting slider allows NFT traders to customize the weighted rarity model. NFT traders can drag the slider to specify the weights of trait rarity and image rarity (\textbf{R2}). The system will rank all NFTs in the collections based on the user-defined weighted rarity model.
The list displays all NFT's basic information. The weighted rarity of each NFT is visualized by stacked bar charts. 
The pricing and other information are shown in text and images to offer NFT traders precise numerical reference. NFT traders can rank basic information to select the most cost-effective NFTs in the collection. The list displays 20 NFTs simultaneously by default. If traders want to browse more NFTs, they can swipe down the list and click the More button at the bottom.

\subsection{NFT Indicator View}
The NFT indicator view (\autoref{fig:system_design}C) is intended to assist NFT traders in analyzing the performance of individual NFTs with the collection (\textbf{R3}). The indicators of NFTs are calculated by the system using the selected period in the collection market view. We use parallel coordinates to display each NFT's indicators to provide an overview of the indicators in the collection. Each axis represents an NFT indicator, and each gray-encoded line represents an NFT by default. When NFT traders select an NFT in the NFT list view, the system will generate a legend encoded with a random color on the right and highlight the line with that color in the NFT indicator view. 
NFT traders can also compare the indications of several NFTs to determine which NFT is superior.
By brushing on the axes in the parallel coordinates, NFT traders can filter the NFTs displayed in the list view to select outliers in the collection, such as inexpensive but rare NFTs.

\subsection{NFT Activity View}
The NFT activity view (\autoref{fig:system_design}D) assists traders in analyzing the selected NFT's transactions. 
The view reveals the pattern of individual NFT transactions in order to prevent NFT traders from purchasing riskier NFTs (\textbf{R4}).
When an NFT is selected in the NFT list view, the NFT activity view will display the NFT's image and traits, the transaction network chart, and the legend of related traders.

The transaction network chart uses a force-directed layout to display the related traders with the selected NFTs and the related traders' other NFTs.
When NFT traders hold dozens of NFTs, the force-guided layout allows traders to easily observe NFT clustering.
To avoid node overlapping, we provide each node with a collision volume equal to its radius.
Traders can also manually adjust the layout to adjust the transaction network chart by dragging and zooming.
The nodes of related traders are depicted in three different shapes (mint \faLegal
, sale
\faTag
, and transfer
\faExchange), which correspond to the various trading behaviors.
The color of each related trader node is encoded with a random color, as shown on the right of the trader legend. The transaction network chart depicts NFTs as circles of equal radius.

\begin{figure}[htb]
    \centering 
    \includegraphics[width=\linewidth]{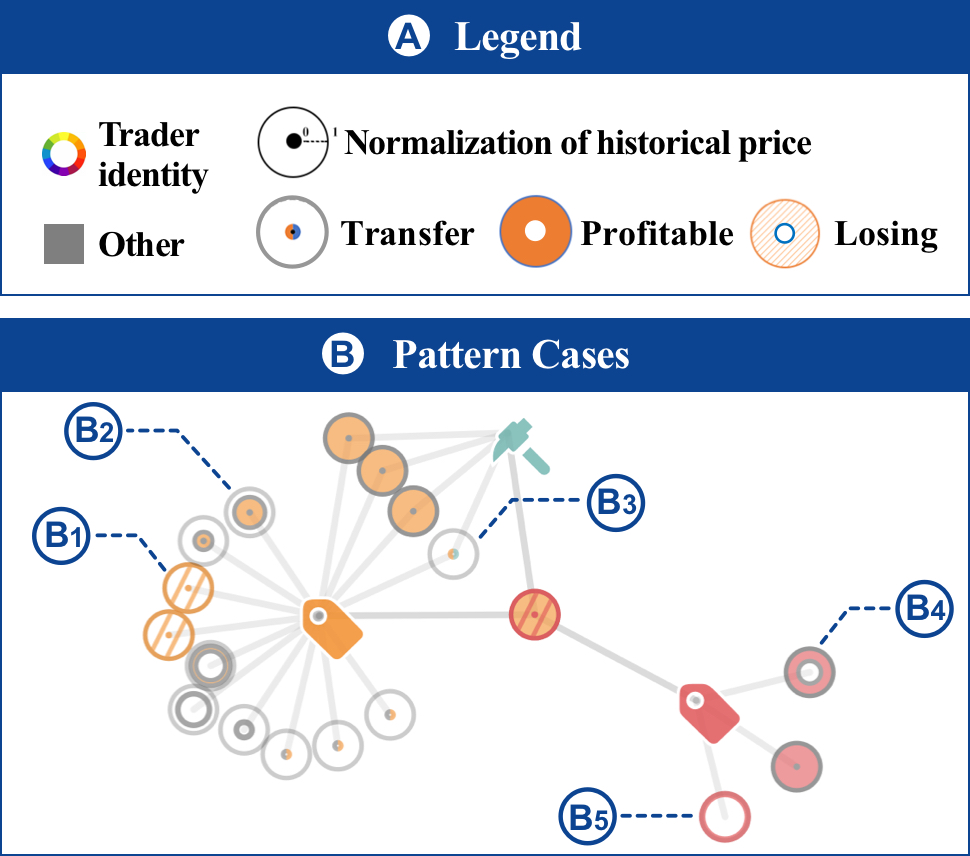}
    \caption{The design of transactions. A) The legend of transaction design. B) The pattern cases about NFTs. 
    B1 indicates that the orange trader purchased the NFT at the highest historical price and sold it at the lowest historical price.
    B2 indicates that the orange trader purchased the NFT at the lowest historical price and sold it at the average historical price.
    B3 indicates that the green trader transfers the NFT to the orange trader.
    B4 indicates that the red trader purchased the NFT at an average historical price and sold it at the highest historical price.
    B5 indicates that the red trader purchased and held the NFT at the highest historical price.
    }
    \label{fig:visual_design}
\end{figure}

The transaction records of NFTs are depicted with various forms of concentric rings within the NFT's circle by design standards \autoref{fig:visual_design}.
The normalized historical transaction prices are mapped to the radius of the concentric circle.
The pre-trade and post-trade prices are mapped to the value of the inner and outer radius of the ring.
The transaction's profitability is encoded with texture.
If the price of NFT has risen since the transaction, it will be encoded with a solid color.
If the price of NFT has fallen after the transaction, it will be encoded with shading. 
The related trader who acquired the NFT in each transaction is mapped to the color of the ring area.
Each concentric circle is drawn according to the sequence of transactions. Therefore, NFT traders can comprehend the transaction order based on the color of the ring area and contours.
Additionally, the transfer does not form a ring in the NFT's circle since it is not profitable.
We use the circle to encode the transfer and color-code them based on the order of their transactions, clockwise around the circle. We use the circle to encode the transfer and color it clockwise based on the order of their transactions.
Traders can observe transaction patterns from different perspectives by dragging nodes. 
When data grows, our design allows for zooming in and out of the view to display more nodes.
Hovering over each NFT circle will display the transaction records in a tooltip, allowing traders to better understand the NFT's transaction records.
Our transaction network chart design summarizes common transaction patterns and enables NFT traders to analyze the Insight behind transaction records more efficiently.

The legend of related traders displays traders' basic information and indicators on the right of the view. The indicators include the holding value, PnL, and the number of activities.
\section{Evaluation}
In this section, we use two case studies and user studies to evaluate \sysName.

\subsection{Case Studies}
The NFT traders who participated in our case studies are our collaborators ($E_1$ and $E_2$).

\subsubsection{Search for Rare and Inexpensive NFTs}
\begin{figure}[htb]
    \centering 
    \includegraphics[width=\linewidth]{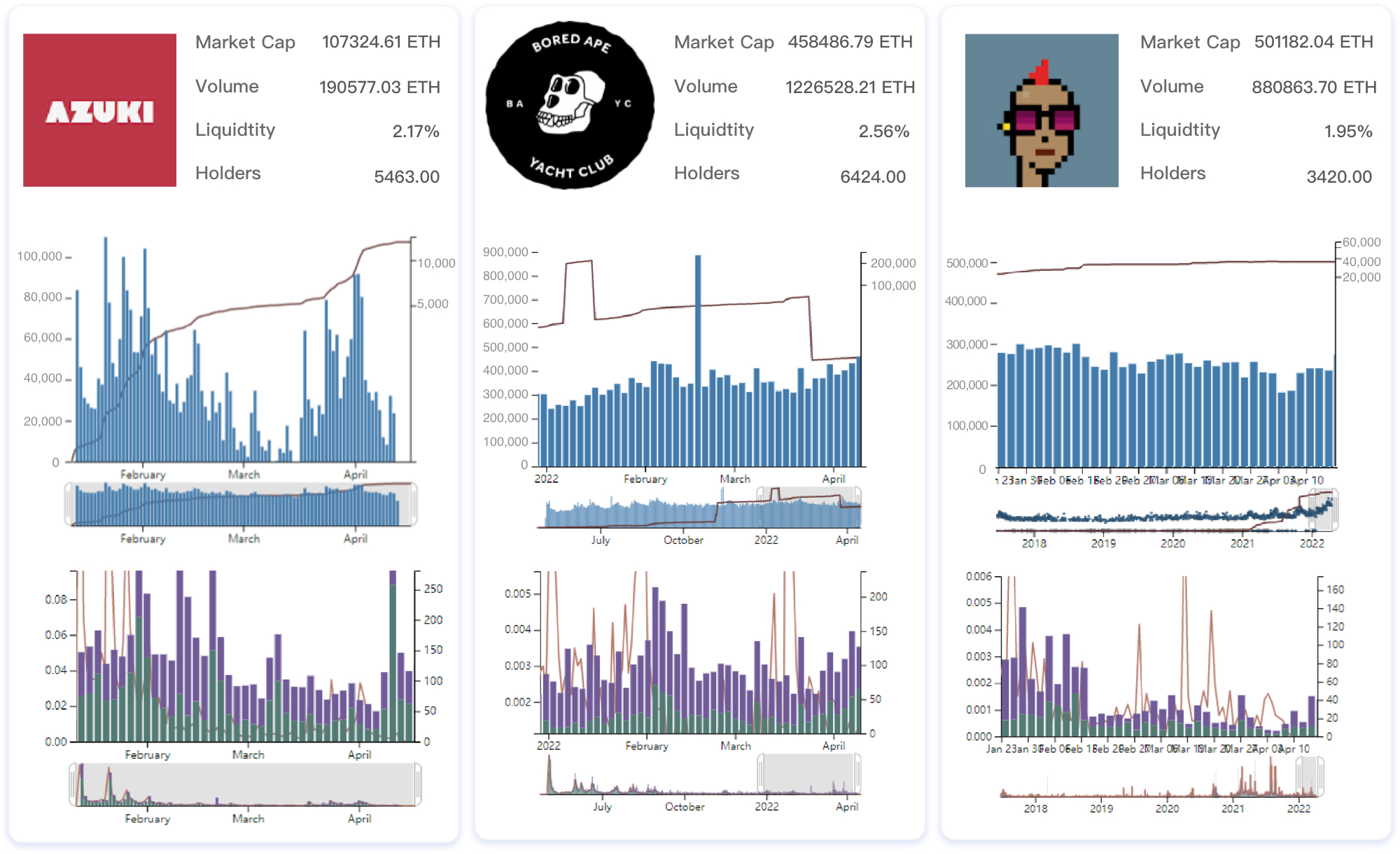}
    \caption{The Azuki collection was established in 2022. It has a small market cap and high liquidity in 2022. Consequently, $E_1$ thought that the Azuki collection was in its early stages of development. 
    The Bored Ape Yacht Club collection was established in 2021. It has a larger market cap, volatility, and less liquidity in 2022. Consequently, $E_1$ thought that the Bored Ape Yacht Club collection was in the middle stages of development.  
    The CryptoPunks collection was established in 2017. It has a more stable market cap and less liquidity in 2022 than the previous two.
    Consequently, $E_1$ thought that the CryptoPunks collection was a mature market.  }
    \label{fig:case1_1}
\end{figure}

$E_1$ has two years of NFT trading experience. He would like to use our system to find a cost-effective NFT. To begin, he compared the collection market evaluation to select a collection. He noticed that the CyptoPunks collection has a more stable market than the other collections (\textbf{R1}), as shown in \autoref{fig:case1_1}. $E_1$ believed that the more mature the market, the more stable the prices of NFTs. Therefore, $E_1$ decided to select an NFT from the collection of CryptoPucks.

Next, $E_1$ was required to select some NFTs that may be worth investing from the NFT list view for in-depth analysis. 
Based on his knowledge of rarity, $E_1$ first assigned equal weight to traits rarity and image rarity (\textbf{R2}). Then, $E_1$ ranked all NFTs based on the custom rarity model and filtered out those that exceeded the price $E_1$ could accept.

\begin{figure}[htb]
    \centering 
    \includegraphics[width=\linewidth]{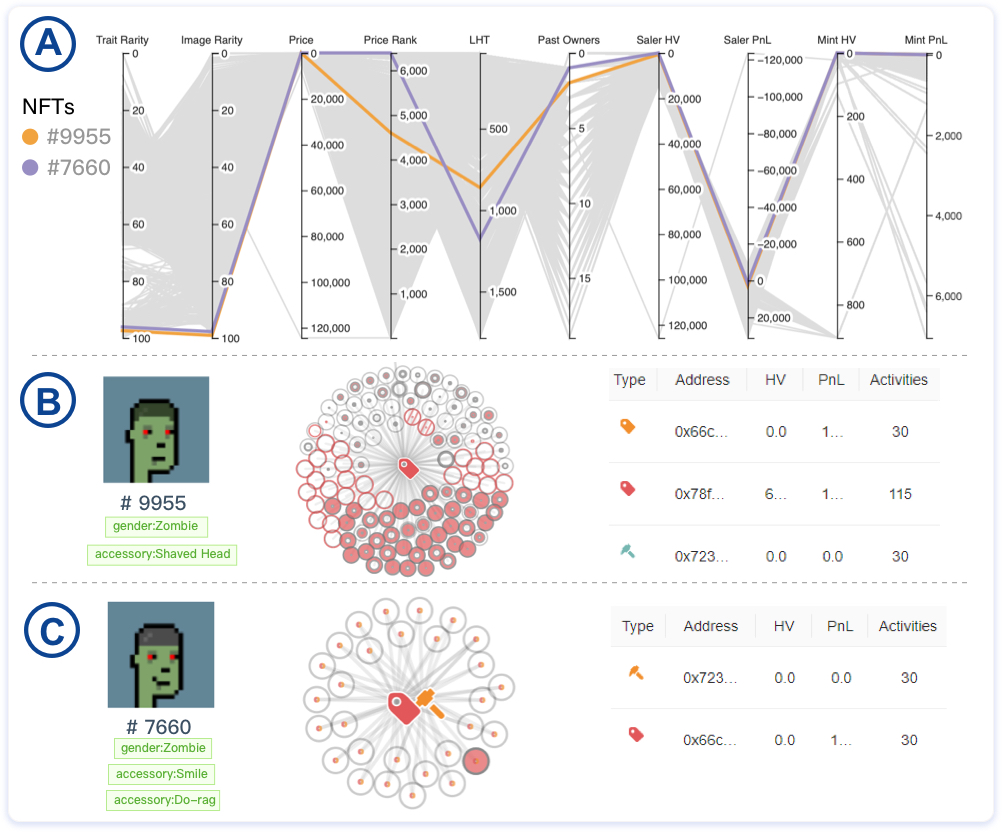}
    \caption{A) Both of these two NFTs are with high image rarity, and low price. B) The owner of NFT \# 9955 has many records of selling. C) The owner of NFT \# 7660 has almost no records of selling NFTs. }
    \label{fig:case1_2}
\end{figure}

Initially, $E_1$ found NFT \#7660. In the NFT indicator view, he could see that NFT \#7660 had a high rarity of both traits and image, and the price of the last transaction was not very high in \autoref{fig:case1_2}A (\textbf{R3}). However, $E_1$ found in the NFT activity view that the current holder of NFT \#7660 had no sales transactions (\textbf{R4}), as shown in \autoref{fig:case1_2}C. The current holder looked like a collector who would not sell his NFTs. Consequently, $E_1$ browsed the NFT list and found another NFT (NFT \#9955), which also had a high rarity and relatively low pricing in \autoref{fig:case1_2}A. Based on the current holder's numerous sales transactions in \autoref{fig:case1_2}B, $E_1$ thought the current holder was an NFT investor who could trade NFT \#9955. Finally, he decided to engage with the current holder of NFT \#9955 regarding the purchase.

\subsubsection{Identify Suspected High-risk NFTs}
\begin{figure*}[htb]
    \centering 
    \includegraphics[width=\linewidth]{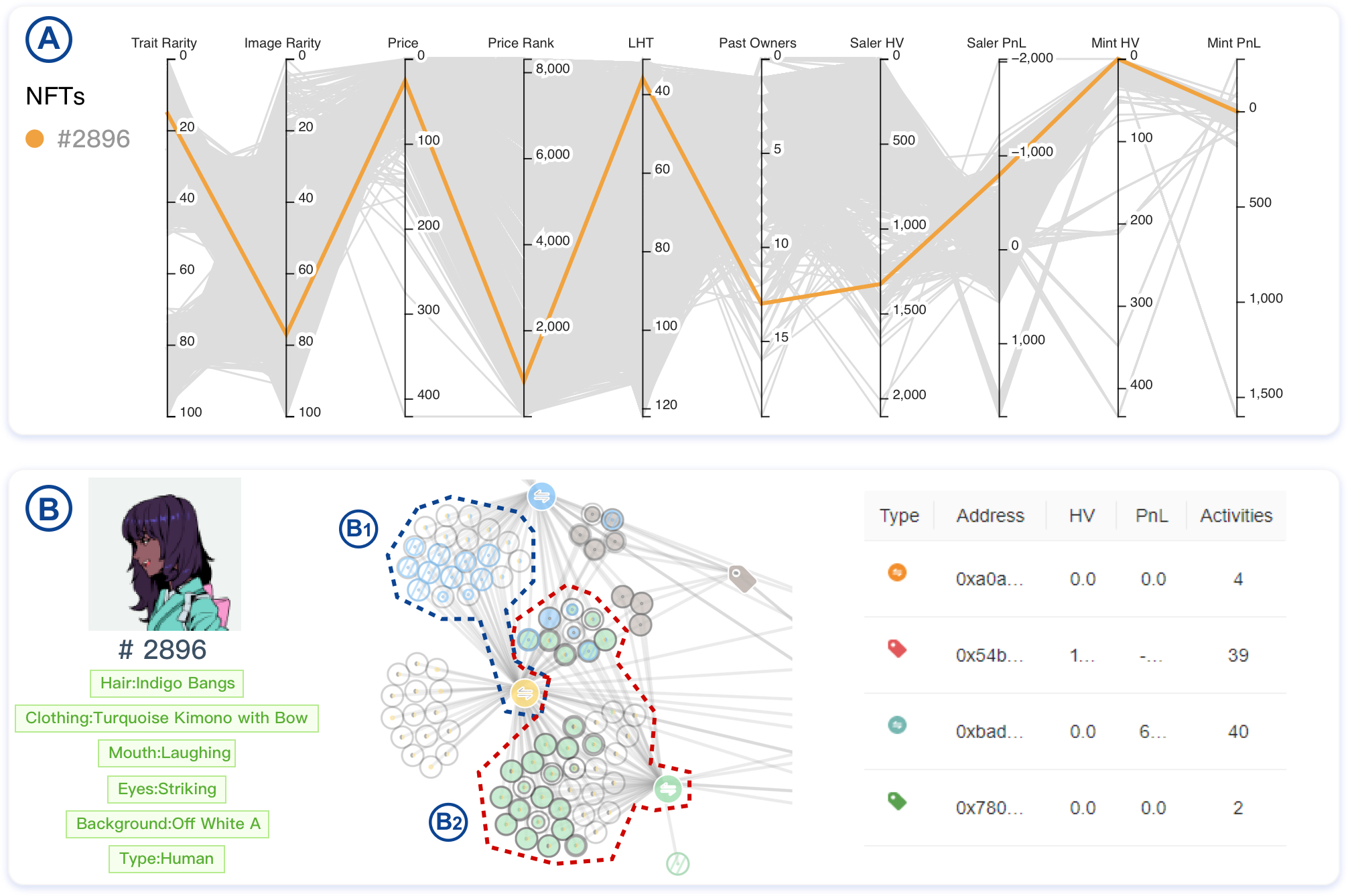}
    \caption{
    The case of identifying suspected high-risk NFTs.
    A) NFT \#2896 has a low trait rarity and high price rank. Its saler PnL is negative.
    B) There are many NFT nodes with blue shading between the blue trader and the yellow trader (B1). Many NFT nodes have half-yellow and overall filled with green between the yellow trader and green trader (B2).
    B1) NFTs with blue shading connect with the yellow trader and the blue trader. These NFTs indicate that the blue trader sold NFTs to the yellow trader for 0 ETH. 
    B2) NFTs with yellow half-centers and overall filled with green connect with the green trader and the yellow trader. These NFTs indicate that the yellow trader transferred these NFTs to the green trader and the green trader sold these NFTs to other traders at a high price.
    }
    \label{fig:case2}
\end{figure*}

$E_2$ has researched NFT for almost three years. He noticed that the prices of NFTs in new collections are relatively low. But there are many risks with these NFTs, such as an inflated price.

Therefore, he wanted to use our system to assess an NFT that he found in Azuki and thought might be risky.
First, $E_2$ searched for NFT \#2896 in the NFT list view and selected it for in-depth analysis.
In the NFT indicator view, he noted that NFT \#2896 had a low trait rarity, a high price rank, and a negative salers' PnL in \autoref{fig:case2}A (\textbf{R3}).
He thought that these indicators' performance was illogical.

Then, he observed the patterns of NFT \#2896 in the NFT activity view (\textbf{R4}).
He found a group of anomalous traders in the transaction network chart, as shown in \autoref{fig:case2}B.
The blue trader collected NFTs continuously at a high price and sold these NFTs to the yellow trader for $0$ ETH in \autoref{fig:case2}$B_1$. The yellow trader transferred these NFTs to the green trader, who sold them at a high price in \autoref{fig:case2}$B_2$.
According to $E_2$'s investment experience, he believed that these transactions were anomalous. 
The indicators' performance of NFT \#2896  was illogical due to these anomalous transactions. Therefore, $E_2$ confirmed that NFT \#2896 was a risky NFT whose price would fall in the future.

\subsection{User Study}
We invited ten participants to evaluate the effectiveness and usability of our system. All of our participants are seasoned NFT investors.

Each participant spent about an hour evaluating our system. To ensure that all participants were familiar with NFTs, we began by introducing the background, data abstract, and indicators. Then, we taught each participant how to use our system through the usage of two use cases.
Next, we asked them to select three cost-effective NFTs using our system within a half-hour and provide an explanation for why these NFTs are cost-effective. After that, we asked them to rate our system on a five-point Likert scale.

In addition, eight
participants with knowledge of rarity were asked to evaluate our rarity model for an additional half-hour.

We randomly selected 100 NFTs from the CryptoPunks collection as our dataset.
Then, we calculated the rarity of these 100 NFTs using the trait rarity model NFTGO, which was currently regarded as the most accurate trait rarity model~\cite{explaining_nft_rarity}, and our image rarity model, respectively.
Next, participants were asked to manually score rarity for the five NFTs with the largest difference in rarity calculated by the two models.
The participants must compare these five NFTs to every other NFT in the dataset. Their judgment is recorded as five levels of difference: similar = 1, slightly similar = 2, poor judgment = 3, slightly different = 4, and very different = 5.
Finally,  the average of the differences is used to determine the participant's rating of the rarity of these five NFTs.

\begin{figure}[htb]
    \centering 
    \includegraphics[width=\linewidth]{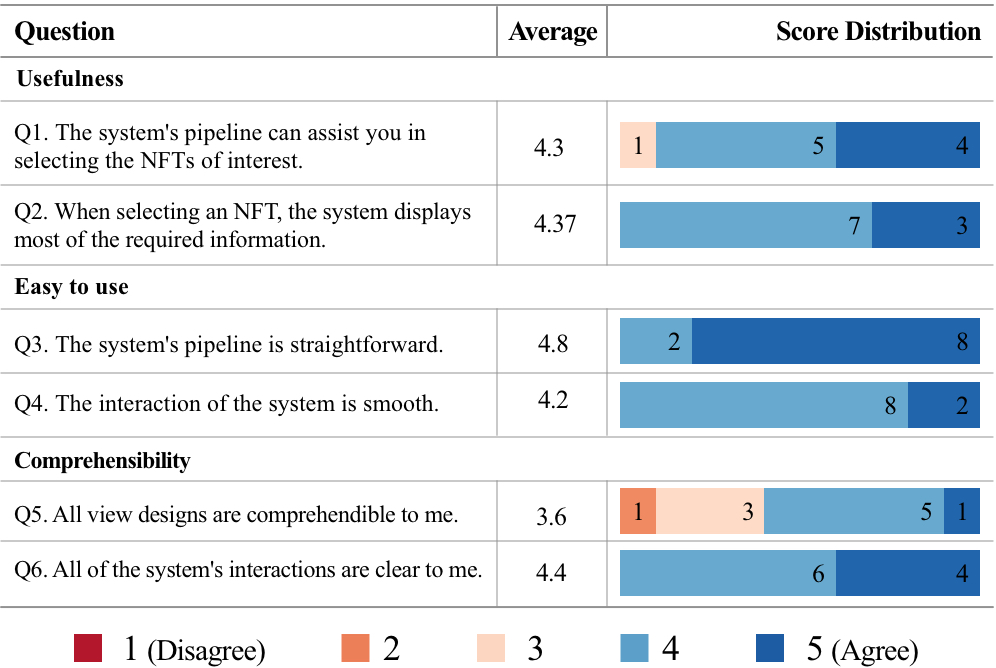}
    \caption{The evaluation of \sysName, including usefulness, whether it is easy to use, and comprehensibility of the visualization.}
    \label{fig:experiment1}
\end{figure}

\begin{figure}[htb]
    \centering 
    \includegraphics[width=\linewidth]{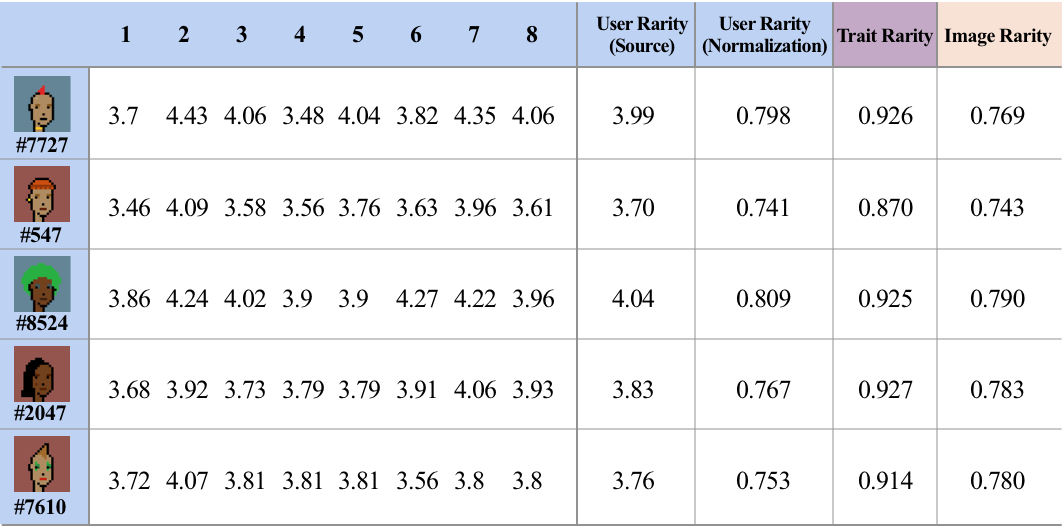}
    \caption{
    The comparison experiment of rarity models. The numbers 1-8 represent the eight participants. User Rarity (Source) is the average rating of these eight participants. The User Rarity (Normalization) transforms the average rating to a value between 0 and 1. The Trait Rarity is calculated by the NFTGO model. The Image Rarity is calculated by our image rarity model.}
    \label{fig:experiment2}
\end{figure}
Seeking iterative optimization, we have conducted two experiments with our participants. 
The final experimental results show that our system received positive feedback from participants, as shown in \autoref{fig:experiment1}. The majority of participants agreed that our system was useful.
$P_2$ and $P_6$ agreed that our system met their requirements for select NFTs. The rich visual view (e.g., collection market and NFT activity view) and interaction (e.g., filtering NFTs in the parallel coordinates and dragging NFT nodes in the transaction network chart) can aid them in selecting the NFTs of interest.
$P_3$ was unsure if our pipeline could help him find the NFTs he desired because he had previously relied solely on NFT traits for selection.
All the participants agreed that our system was easy to use. $P_4$ said that our system provides a straightforward analysis pipeline, which corresponds to his experience.
However, some participants felt that some of the visualizations in our system are hard to understand. 
$P_4$, $P_9$, and $P_{10}$ pointed out that the design of the transactions was too complex to understand directly. 
On occasion, they had to investigate historical transactions based on the tooltips that required them to mouse over the design. 
Nevertheless, $P_3$ and $P_8$ agreed that the design of the transaction network chart was useful in summarizing the transaction patterns.
$P_3$ said, ``The activity view illustrates the related trader's transaction patterns in a novel manner. Before this user study, I focused on the performance of NFT itself. Now, I see that historical traders have a significant impact on NFT's performance.''
$P_7$ commented that although some transactions were difficult to understand without tooltips, it would be useful for analyzing patterns if familiar with the common transactions.

\autoref{fig:experiment2} depicts the evaluation of the rarity model. The results indicate that participant perceptions of NFT rarity resemble image rarity more than trait rarity. However, when we communicated with the participants, we learned that they held varying opinions on the rarity model. 
$P_3$, $P_5$, $P_7$, and $P_8$ indicated that image rarity corresponded to their conceptions of rarity, and they believed that image rarity might be more significant than trait rarity.
$P_6$, $P_9$, and $P_{10}$ pointed out that image rarity cannot replace trait rarity.
$P_6$ stated, ``If an NFT has a trait that no other NFT has, it is one of the rarest NFTs, even if it is similar to other NFTs in many places.'' So he believed that the image rarity was valuable but far less important than trait rarity.
$P_2$ clarified that although it is difficult to determine which rarity is more important, image rarity can undoubtedly assist NFT traders in locating rare NFTs, and the potential value of some NFTs will likely be reassessed.


\section{Discussion}
We presented two case studies to demonstrate the usefulness of our system. And in our user studies, we received a lot of positive feedback from the participants. They approved that our system and rarity model could help NFT traders analyze the performance of NFTs efficiently. 
What's more, our system can also help market regulators analyze risks in the NFT market. Currently, there are a lot of abnormal transactions in the NFT market. Some countries are cracking down on illegal NFT transactions to reduce crime~\cite{crime1,crime2}. Our system can visualize traders' transaction patterns to help regulators discover possible anomalous transactions. However, there are still two limitations to improve in our system.

\textbf{Design Scalability}
NFT transactions are critical data that can reveal a variety of interesting insights, such as risky NFTs, risky NFT traders, NFT trader profitability, and so on.
Current NFT information service platforms only display NFT transactions as a list, which makes it difficult for NFT traders to discern the insights underlying the transactions.
To address this issue, we design new designs on the transaction network chart to illustrate the transaction patterns underlying individual NFTs and to assist NFT traders in analyzing insights effectively.
Through case study and use study, we demonstrated that our designs enable users to reveal insights from NFT transactions.
However, as the amount of transaction records grows, our design struggles to visually represent some complex transaction patterns.
NFT traders necessitate hovering over the NFT nodes to comprehend the transaction records. In the future, we will research how to use visualization to address this problem.

\textbf{System Scalability}
Our system includes a variety of NFT indicators and a comprehensive presentation of NFT performance, but different traders have different preferences.
$P_3$ and $P_5$ pointed out that if our system could present customized indicators, their analysis would be more efficient. For example, $P_3$ considered the price rank to be meaningless. If our system allowed for the deletion or addition of indicators in the NFT indicator view, it would be easier for him to analyze the correlation between NFT indicators.
$P_5$ noted that the HV and PnL of transfer traders in NFTs are useful indicators. He hoped that our system would allow adding these indicators.
\section{Conclusion}
In this research, we present a visual analysis approach for effectively assessing the performance of NFTs. Moreover, we present a new rarity model that can quantify NFTs with images and traits. Through two case studies and user studies, the effectiveness and usefulness of our system are evaluated.

In the future, we will open-source the system. All NFT traders can use \sysName{} to analyze and assess the performance of individual NFTs. Additionally, we will attempt to solve the issue of how to visualize some complex transactions in numerous transactions intuitively. We intend to use visualization to facilitate more FinTech applications, such as NFTs.

\acknowledgments{
We wish to thank all the anonymous reviewers for their thorough and constructive comments. We also thank the participants for their time and efforts. This work was supported by the National Natural Science Foundation of China (No. 62132017, 62202244).
}

\bibliographystyle{abbrv-doi}

\bibliography{template}

\begin{thebibliography}{10}

\bibitem{azuki}
{AZUKI}.
\newblock \url{https://www.azuki.com/}.
\newblock Accessed May, 2022.

\bibitem{bayc}
Bored ape yacht club.
\newblock \url{https://boredapeyachtclub.com/}.
\newblock Accessed May, 2022.

\bibitem{crime2}
{China’s banking association proposes restrictions on NFTs}.
\newblock
  \url{https://techcrunch.com/2022/04/13/chinas-banking-association-proposes-restrictions-on-nfts/}.
\newblock Accessed Oct, 2022.

\bibitem{crime1}
{China’s first court ruling on NFT art theft holds marketplace accountable}.
\newblock
  \url{https://www.scmp.com/tech/tech-trends/article/3175457/chinas-first-court-ruling-nft-art-theft-holds-marketplace}.
\newblock Accessed Oct, 2022.

\bibitem{cryptoart}
Cryptoart.
\newblock \url{https://cryptoart.io/}.
\newblock Accessed May, 2022.

\bibitem{cryptopunks}
Cryptopunks.
\newblock \url{https://cryptopunks.app/}.
\newblock Accessed May, 2022.

\bibitem{element}
Element.
\newblock \url{https://element.market/}.
\newblock Accessed May, 2022.

\bibitem{explaining_nft_rarity}
{Explaining NFT rarity score models}.
\newblock
  \url{https://www.publish0x.com/navid-ladani/explaining-nft-rarity-score-models-xmmorpr}.
\newblock Accessed Oct, 2022.

\bibitem{jdsupra20220428}
{Into the metaverse: M\&A in the NFT market in 2022}.
\newblock
  \url{https://www.jdsupra.com/legalnews/into-the-metaverse-m-a-in-the-nft-8112977/}.
\newblock Accessed Oct, 2022.

\bibitem{nftgo}
{NFTGO}.
\newblock \url{https://nftgo.io/}.
\newblock Accessed May, 2022.

\bibitem{nftgo_rarity}
{NFTGO rarity model}.
\newblock \url{https://docs.nftgo.io/docs/nftgo-rariry-model}.
\newblock Accessed Oct, 2022.

\bibitem{OKLink}
{OKLink}.
\newblock \url{https://www.oklink.com/en/nft}.
\newblock Accessed May, 2022.

\bibitem{opensea}
Opensea.
\newblock \url{https://opensea.io/}.
\newblock Accessed May, 2022.

\bibitem{rarity_tools_model}
{Ranking rarity: Understanding rarity calculation methods}.
\newblock
  https://raritytools.medium.com/ranking-rarity-understanding-rarity-calculation-
  methods-86ceaeb9b98c.
\newblock Accessed Oct, 2022.

\bibitem{rarity.tools}
Rarity.tools.
\newblock \url{https://rarity.tools/}.
\newblock Accessed May, 2022.

\bibitem{fortune20220505}
{The NFT market is probably not as down as you think it is}.
\newblock
  \url{https://fortune.com/2022/05/05/nft-market-sales-not-flatlining-data-shows/}.
\newblock Accessed Oct, 2022.

\bibitem{ante2021non}
L.~Ante.
\newblock {Non-fungible token (NFT) markets on the Ethereum blockchain:
  Temporal development, cointegration and interrelations}.
\newblock {\em Available at SSRN}, 2021.

\bibitem{ante2021non2}
L.~Ante.
\newblock {The non-fungible token (NFT) market and its relationship with
  Bitcoin and Ethereum}.
\newblock {\em Available at SSRN}, 2021.

\bibitem{basole2017visualization}
R.~C. Basole, M.~A. Bellamy, and H.~Park.
\newblock Visualization of innovation in global supply chain networks.
\newblock {\em Decision Sciences}, 48(2):288--306, 2017.

\bibitem{beis1997shape}
J.~S. Beis and D.~G. Lowe.
\newblock Shape indexing using approximate nearest-neighbour search in
  high-dimensional spaces.
\newblock In {\em Proceedings of IEEE Computer Society Conference on Computer
  Vision and Pattern Recognition}, pp. 1000--1006, 1997.

\bibitem{bostock2011d3}
M.~Bostock, V.~Ogievetsky, and J.~Heer.
\newblock D$^3$ data-driven documents.
\newblock {\em IEEE Transactions on Visualization and Computer Graphics},
  17(12):2301--2309, 2011.

\bibitem{chohan2021non}
U.~W. Chohan.
\newblock Non-fungible tokens: Blockchains, scarcity, and value.
\newblock {\em Critical Blockchain Research Initiative (CBRI) Working Papers},
  2021.

\bibitem{dowling2022fertile}
M.~Dowling.
\newblock Fertile land: Pricing non-fungible tokens.
\newblock {\em Finance Research Letters}, 44:102096, 2022.

\bibitem{dowling2022non}
M.~Dowling.
\newblock Is non-fungible token pricing driven by cryptocurrencies?
\newblock {\em Finance Research Letters}, 44:102097, 2022.

\bibitem{dwyer2002visualising}
T.~Dwyer and P.~Eades.
\newblock Visualising a fund manager flow graph with columns and worms.
\newblock In {\em Proceedings of Proceedings Sixth International Conference on
  Information Visualisation}, pp. 147--152, 2002.

\bibitem{dwyer2004visualising}
T.~Dwyer and D.~R. Gallagher.
\newblock Visualising changes in fund manager holdings in two and a
  half-dimensions.
\newblock {\em Information Visualization}, 3(4):227--244, 2004.

\bibitem{fleder2015bitcoin}
M.~Fleder, M.~S. Kester, and S.~Pillai.
\newblock Bitcoin transaction graph analysis.
\newblock {\em Arxiv Preprint arXiv:1502.01657}, 2015.

\bibitem{franceschet2021hits}
M.~Franceschet.
\newblock Hits hits art.
\newblock {\em Blockchain: Research and Applications}, 2(4):100038, 2021.

\bibitem{franceschet2021sentiment}
M.~Franceschet.
\newblock The sentiment of crypto art.
\newblock {\em Proceedings http://ceur-ws. Org ISSN}, 1613:0073, 2021.

\bibitem{goh2013riskvis}
R.~S.~M. Goh, Z.~Wang, X.~Yin, X.~Fu, L.~Ponnambalam, S.~Lu, and X.~Li.
\newblock {RiskVis: Supply chain visualization with risk management and
  real-time monitoring}.
\newblock In {\em Proceedings of IEEE International Conference on Automation
  Science and Engineering (CASE)}, pp. 207--212, 2013.

\bibitem{kapoor2022tweetboost}
A.~Kapoor, D.~Guhathakurta, M.~Mathur, R.~Yadav, M.~Gupta, and P.~Kumaraguru.
\newblock {TweetBoost: Influence of social media on NFT valuation}.
\newblock {\em Arxiv Preprint arXiv:2201.08373}, 2022.

\bibitem{keim2006spectral}
D.~A. Keim, T.~Nietzschmann, N.~Schelwies, J.~Schneidewind, T.~Schreck, and
  H.~Ziegler.
\newblock A spectral visualization system for analyzing financial time series
  data.
\newblock In {\em Proceedings of Eurographics/IEEE TCVG Symposium on
  Visualization}, pp. 195--202, 2006.

\bibitem{kleinberg1999authoritative}
J.~M. Kleinberg.
\newblock Authoritative sources in a hyperlinked environment.
\newblock {\em Journal of the ACM (JACM)}, 46(5):604--632, 1999.

\bibitem{ko2016survey}
S.~Ko, I.~Cho, S.~Afzal, C.~Yau, J.~Chae, A.~Malik, K.~Beck, Y.~Jang,
  W.~Ribarsky, and D.~S. Ebert.
\newblock A survey on visual analysis approaches for financial data.
\newblock In {\em Proceedings of Computer Graphics Forum}, pp. 599--617, 2016.

\bibitem{lipton2021automated}
A.~Lipton and A.~Sepp.
\newblock Automated market-making for fiat currencies.
\newblock {\em Available at SSRN}, 2021.

\bibitem{lowe1999object}
D.~G. Lowe.
\newblock Object recognition from local scale-invariant features.
\newblock In {\em Proceedings of the Seventh IEEE International Conference on
  Computer Vision}, pp. 1150--1157, 1999.

\bibitem{lowe2004distinctive}
D.~G. Lowe.
\newblock Distinctive image features from scale-invariant keypoints.
\newblock {\em International Journal of Computer Vision}, 60(2):91--110, 2004.

\bibitem{maouchi2021understanding}
Y.~Maouchi, L.~Charfeddine, and G.~El~Montasser.
\newblock {Understanding digital bubbles amidst the COVID-19 pandemic: Evidence
  from DeFi and NFTs}.
\newblock {\em Finance Research Letters}, p. 102584, 2021.

\bibitem{mcginn2016visualizing}
D.~McGinn, D.~Birch, D.~Akroyd, M.~Molina-Solana, Y.~Guo, and W.~J.
  Knottenbelt.
\newblock Visualizing dynamic bitcoin transaction patterns.
\newblock {\em Big Data}, 4(2):109--119, 2016.

\bibitem{nadini2021mapping}
M.~Nadini, L.~Alessandretti, F.~Di~Giacinto, M.~Martino, L.~M. Aiello, and
  A.~Baronchelli.
\newblock {Mapping the NFT revolution: market trends, trade networks, and
  visual features}.
\newblock {\em Scientific Reports}, 11(1):1--11, 2021.

\bibitem{pantz2013pnl}
J.~Pantz.
\newblock Pnl prediction under extreme scenarios.
\newblock {\em Available at SSRN}, 2013.

\bibitem{parham2022non}
A.~Parham and C.~Breitinger.
\newblock Non-fungible tokens: Promise or peril?
\newblock {\em Arxiv Preprint arXiv:2202.06354}, 2022.

\bibitem{pelechrinis2022spotting}
K.~Pelechrinis, X.~Liu, P.~Krishnamurthy, and A.~Babay.
\newblock {Spotting anomalous trades in NFT markets: The case of NBA topshot}.
\newblock {\em Arxiv Preprint arXiv:2202.04013}, 2022.

\bibitem{phillips202110}
D.~Phillips.
\newblock {The 10 most expensive NFTs ever sold}, 2021.

\bibitem{pinto2022nft}
C.~Pinto-Gutiérrez, S.~Gaitán, D.~Jaramillo, and S.~Velasquez.
\newblock {The NFT hype: What draws attention to non-fungible tokens?}
\newblock {\em Mathematics}, 10(3):335, 2022.

\bibitem{sharma2022s}
T.~Sharma, Z.~Zhou, Y.~Huang, and Y.~Wang.
\newblock {``It's a blessing and a curse'': Unpacking creators' practices with
  non-fungible tokens (NFTs) and their communities}.
\newblock {\em Arxiv Preprint arXiv:2201.13233}, 2022.

\bibitem{shneiderman1992tree}
B.~Shneiderman.
\newblock Tree visualization with tree-maps: 2-d space-filling approach.
\newblock {\em ACM Transactions on Graphics (TOG)}, 11(1):92--99, 1992.

\bibitem{siddiqui2008supply}
A.~Siddiqui, M.~Khan, and S.~Akhtar.
\newblock Supply chain simulator: A scenario-based educational tool to enhance
  student learning.
\newblock {\em Computers \& Education}, 51(1):252--261, 2008.

\bibitem{singh2021distributed}
J.~Singh and P.~Singh.
\newblock Distributed ownership model for non-fungible tokens.
\newblock {\em Smart and Sustainable Intelligent Systems}, pp. 307--321, 2021.

\bibitem{smeulders2005interactive}
R.~Smeulders and A.~Heijs.
\newblock Interactive visualization of high dimensional marketing data in the
  financial industry.
\newblock In {\em Proceedings of Ninth International Conference on Information
  Visualisation (IV'05)}, pp. 814--817, 2005.

\bibitem{trautman2021virtual}
L.~J. Trautman.
\newblock Virtual art and non-fungible tokens.
\newblock {\em Available at SSRN}, 2021.

\bibitem{umar2022covid}
Z.~Umar, M.~Gubareva, T.~Teplova, and D.~K. Tran.
\newblock {COVID-19 impact on NFTs and major asset classes interrelations:
  Insights from the wavelet coherence analysis}.
\newblock {\em Finance Research Letters}, p. 102725, 2022.

\bibitem{vasan2022quantifying}
K.~Vasan, M.~Janosov, and A.-L. Barabási.
\newblock {Quantifying NFT-driven networks in crypto art}.
\newblock {\em Scientific Reports}, 12(1):1--11, 2022.

\bibitem{vidal2022new}
D.~Vidal-Tomás.
\newblock {The new crypto niche: NFTs, play-to-earn, and metaverse tokens}.
\newblock {\em Finance Research Letters}, p. 102742, 2022.

\bibitem{wang2021non}
Q.~Wang, R.~Li, Q.~Wang, and S.~Chen.
\newblock {Non-fungible token (NFT): Overview, evaluation, opportunities and
  challenges}.
\newblock {\em Arxiv Preprint arXiv:2105.07447}, 2021.

\bibitem{wattenberg1999visualizing}
M.~Wattenberg.
\newblock Visualizing the stock market.
\newblock In {\em Proceedings of CHI'99 Extended Abstracts on Human Factors in
  Computing Systems}, pp. 188--189, 1999.

\bibitem{whitaker2019art}
A.~Whitaker.
\newblock Art and blockchain: A primer, history, and taxonomy of blockchain use
  cases in the arts.
\newblock {\em Artivate}, 8(2):21--46, 2019.

\bibitem{xia2020supoolvisor}
J.-z. Xia, Y.-h. Zhang, H.~Ye, Y.~Wang, G.~Jiang, Y.~Zhao, C.~Xie, X.-y. Kui,
  S.-h. Liao, and W.-p. Wang.
\newblock Supoolvisor: a visual analytics system for mining pool surveillance.
\newblock {\em Frontiers of Information Technology \& Electronic Engineering},
  21(4):507--523, 2020.

\bibitem{yue2019sportfolio}
X.~Yue, J.~Bai, Q.~Liu, Y.~Tang, A.~Puri, K.~Li, and H.~Qu.
\newblock {Sportfolio: Stratified visual analysis of stock portfolios}.
\newblock {\em {IEEE Transactions on Visualization and Computer Graphics}},
  26(1):601--610, 2019.

\bibitem{yue2021iquant}
X.~Yue, Q.~Gu, D.~Wang, H.~Qu, and Y.~Wang.
\newblock iquant: Interactive quantitative investment using sparse regression
  factors.
\newblock In {\em Proceedings of Computer Graphics Forum}, pp. 189--200, 2021.

\bibitem{yue2018bitextract}
X.~Yue, X.~Shu, X.~Zhu, X.~Du, Z.~Yu, D.~Papadopoulos, and S.~Liu.
\newblock Bitextract: Interactive visualization for extracting bitcoin exchange
  intelligence.
\newblock {\em {IEEE Transactions on Visualization and Computer Graphics}},
  25(1):162--171, 2018.

\end{thebibliography}
\end{document}